\documentclass{article}

\usepackage{arxiv}
\usepackage[utf8]{inputenc} % allow utf-8 input
\usepackage[T1]{fontenc}    % use 8-bit T1 fonts
\usepackage{hyperref}       % hyperlinks
\usepackage{url}            % simple URL typesetting
\usepackage{booktabs}       % professional-quality tables
\usepackage{amsfonts}       % blackboard math symbols
\usepackage{nicefrac}       % compact symbols for 1/2, etc.
\usepackage{microtype}      % microtypography
\usepackage{lipsum}
\usepackage{graphicx}
\graphicspath{ {./images/} }

\title{Achievements and Challenges in Explaining Deep Learning based Computer-Aided Diagnosis Systems}

\author{
 Adriano Lucieri\textsuperscript{*} \\
  Department of Computer Science\\
  Technical University of Kaiserslautern\\
  67663 Kaiserslautern, Germany \\
  Smart Data and Knowledge Services \\
  German Research Center for AI GmbH (DFKI) \\
  67663 Kaiserslautern, Germany \\
  \texttt{adriano.lucieri@dfki.de} \\
   \And
Muhammad Naseer Bajwa\textsuperscript{*} \\
  Department of Computer Science\\
  Technical University of Kaiserslautern\\
  67663 Kaiserslautern, Germany \\
  Smart Data and Knowledge Services \\
  German Research Center for AI GmbH (DFKI) \\
  67663 Kaiserslautern, Germany \\
  \texttt{naseer.bajwa@dfki.de} \\
  \And
Andreas Dengel \\
  Department of Computer Science\\
  Technical University of Kaiserslautern\\
  67663 Kaiserslautern, Germany \\
  Smart Data and Knowledge Services \\
  German Research Center for AI GmbH (DFKI) \\
  67663 Kaiserslautern, Germany \\
  \texttt{andreas.dengel@dfki.de} \\
    \And
Sheraz Ahmed \\
  Smart Data and Knowledge Services \\
  German Research Center for AI GmbH (DFKI) \\
  67663 Kaiserslautern, Germany \\
  \texttt{sheraz.ahmed@dfki.de} \\
}

\begin{document}
\thanks{Authors contributed equally}
\thanks{This is English translation of a German book chapter to appear in Springer Nature 'KI Im Gesundheitswesen'.}
\maketitle
\begin{abstract}
Remarkable success of modern image-based AI methods and the resulting interest in their applications in critical decision-making processes has led to a surge in efforts to make such intelligent systems transparent and explainable. The need for explainable AI does not stem only from ethical and moral grounds but also from stricter legislation around the world mandating clear and justifiable explanations of any decision taken or assisted by AI. Especially in the medical context where Computer-Aided Diagnosis can have a direct influence on the treatment and well-being of patients, transparency is of utmost importance for safe transition from lab research to real world clinical practice. This paper provides a comprehensive overview of current state-of-the-art in explaining and interpreting Deep Learning based algorithms in applications of medical research and diagnosis of diseases. We discuss early achievements in development of explainable AI for validation of known disease criteria, exploration of new potential biomarkers, as well as methods for the subsequent correction of AI models. Various explanation methods like visual, textual, post-hoc, ante-hoc, local and global have been thoroughly and critically analyzed. Subsequently, we also highlight some of the remaining challenges that stand in the way of practical applications of AI as a clinical decision support tool and provide recommendations for the direction of future research.
\end{abstract}


\section{Introduction}
\label{sec:intro}
Artificial Intelligence (AI) based methods to support medical professionals were developed as early as 1970s~\cite{buchanan1969heuristic, shortliffe1974mycin}. However, hardly any of these methods ever found practical applications in practical clinical environment, which was partly due to the technological status of computer systems at that time and emerging ethical and legal concerns. The exponential growth in computing power and digital data volume in the past decades, coupled with maturing of data-driven algorithms like Deep Neural Networks (DNNs), led to a number of groundbreaking successes in areas such as image recognition~\cite{krizhevsky2017imagenet} and speech recognition~\cite{hinton2012deep}. From the outset of the success story of image-based Deep Learning (DL) models, like Convolutional Neural Networks (CNNs), an interest in their application in medicine developed immediately. The supremacy of modern DL-based solutions in the field of automated diagnosis of medical images over conventional image processing methods quickly became apparent~\cite{cruz2013deep, gulshan2016development}. Shortly afterwards, early comparative studies reported the superiority of such self-learning algorithms over human experts in fields such as dermatology~\cite{esteva2017dermatologist, brinker2019deep}, ophthalmology~\cite{abramoff2018pivotal, brown2018fully} and radiology~\cite{rajpurkar2018deep}. In spite of enormous technical and political progress, most ethical concerns about the fairness and transparency of such algorithms still remain unresolved. Unlike earlier rule-based or linear AI models, increasingly complex DL methods often act as black-boxes which do not offer semantically traceable decision processes. Moreover, legal requirements such as European General Data Protection Regulation (GDPR)~\cite{GDPR2018}, which came into effect in 2018, stress compulsory obligations regarding the transparency of automated decision-making processes towards affected individuals.

These factors, among others, contributed to current trend in the field of AI-based diagnosis to move towards Computer-Aided Diagnosis (CAD) and so called "Augmented Doctor"~\cite{ama2018}. Advantages such as speed, objectiveness, and thoroughness of AI methods are being used, for example, as an assistance system to point out relevant disease indicators to doctors, give diagnosis suggestions and to present similar past cases for comparison. This approach of clinical application benefits from being a Human-In-The-Loop (HITL) hybrid keeping the clinical experts in control of the process~\cite{singh2020explainable}. This HITL model is similar to driving aids like adaptive cruise control or lane keep assistance in automobiles where human driver retains control and bears responsibility for the final decisions but with a reduced workload and an added safety net. In healthcare, this implies that the indicators and explanations presented by AI move to the centre of attention and that the ultimate diagnosis will be of less relevance as the doctor will always make the final decision. The ability of a diagnostic assistance system to explain its decision path has been considered as most important characteristic by health professionals since long~\cite{teach1981analysis}. This and more general demand for justifications of automated high-stakes decisions led to an enormous increase in research in the field of explainable AI (xAI)~\cite{arrieta2020explainable}. The claim of this article is not to provide an extensive list of previous works and efforts to explain DNNs in medical problems nor to define the notion of explainability and interpretability. Interested readers will find broad overviews of the applications of xAI methods in general medical problems in~\cite{tjoa2020quantifying, stiglic2020interpretability} as well as special works on the topics of medical image analysis~\cite{singh2020explainable} and digital pathology~\cite{vilone2020explainable}. Also, there is no collectively agreed upon xAI vocabulary, but there have been numerous efforts aimed at defining key terms and taxonomies~\cite{gilpin2018explaining, lipton2018mythos,arrieta2020explainable}. In this work, we follow the definitions of explainability and interpretability provided in~\cite{montavon2018methods}.

In this article we take an application-oriented look at the current achievements in the field of the xAI methods to distil remaining challenges in the way of its practical application. We show that in many cases the technical bases for transparent AI applications in medicine already exist and that the missing piece of the puzzle is often extensive cooperation of AI developers and medical domain experts. We also caution readers to not overconfidently rely on any xAI method but be aware of the difficulties in objective evaluation of fidelity and quality of explanations. It is emphasised that AI developers should pay special attention to application-specific characteristics such as human-centricity, domain-orientation, diversity and completeness of explanations when developing xAI methods in future. The following section provides a short overview of some existing xAI methods. In section \ref{sec:achievements} we present current achievements in application of explainable DL methods to medical problems. This includes explanation-based improvement of algorithms, validation of DNNs decision processes, discovery of previously unknown disease criteria, as well as proposed DL-based xAI frameworks for CAD. Section \ref{sec:challenges} sheds a critical light some of the most strenuous challenges that bar the way to progress. Finally, the findings are concluded in section \ref{sec:conclusion}.

\section{Overview of Common xAI Methods}
\label{sec:overview}

\begin{figure}[t!]
    \centering
    \includegraphics[width=0.9\textwidth]{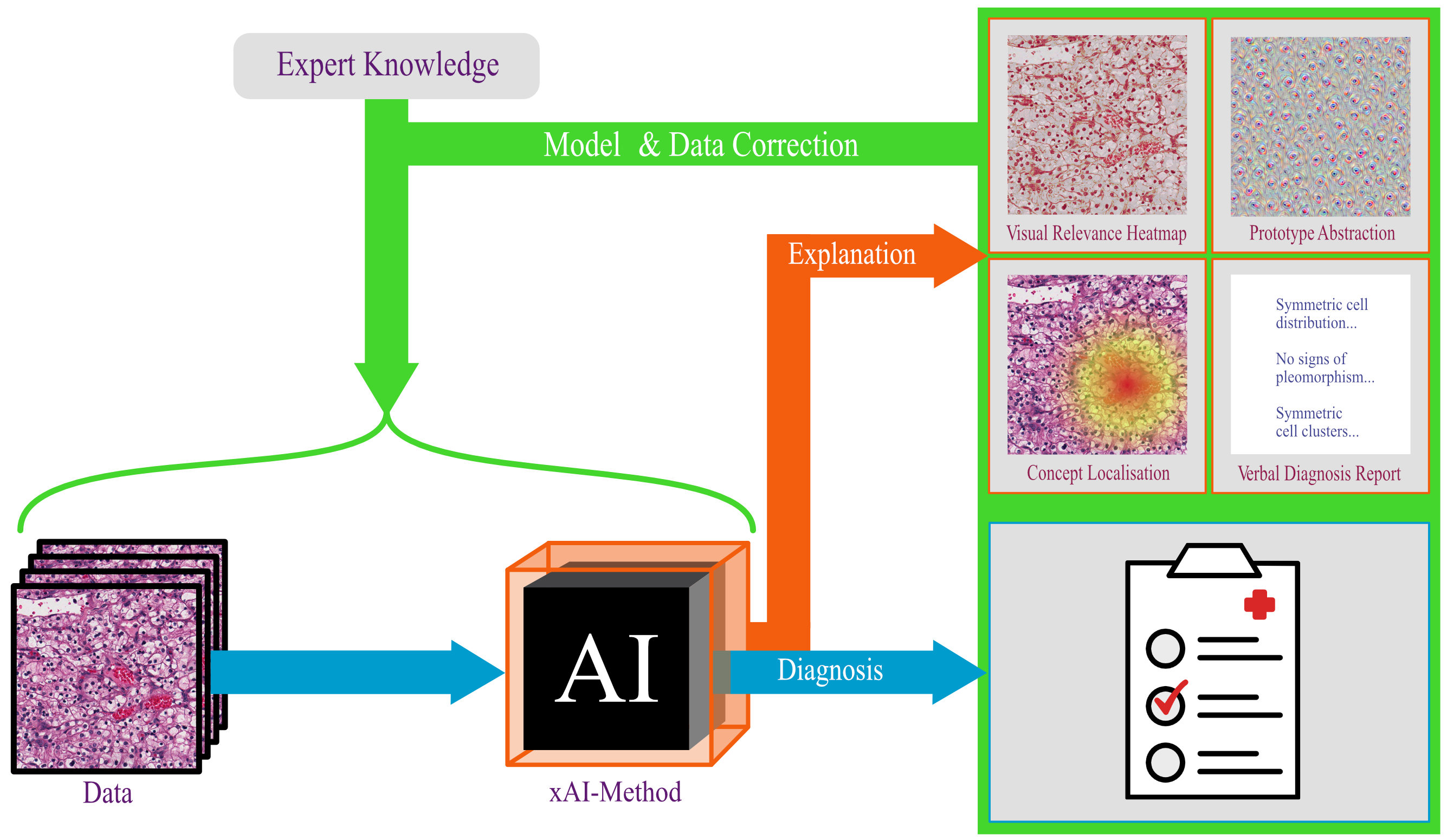}
    \caption{Topology of the xAI process with optional model and data correction as well as taxonomy of common and relevant xAI methods in medical image analysis.}
    \label{fig:topology}
\end{figure}

Methods explaining the decision-making process of DNNs exist in a variety of forms. Not only the derivation of the explanations differs, but also the way it is communicated to the user. There are a number of taxonomies available in literature to differentiate these methods. An important distinction for AI users, for example, is made between post-hoc and ante-hoc methods. Methods which can explain the decision of a black-box model afterwards are called post-hoc (lat.: after-this event) methods. Ante-hoc (lat.: before-this event) methods, on the other hand, are already interpretable due to their architecture. Since this ante-hoc explanations are usually achieved by architectural or conceptual restrictions in the learning process that limits modelling capacity, such inherently interpretable models are often thought to be inferior to their unrestricted conspecifics in terms of final model performance. However, this effect can sometimes be mitigated by pre-processing raw data with noisy features into meaningfully structured representations~\cite{rudin2019stop}. Another important distinction is made between explanation methods that attempt to explain a classifier’s decision process on a global scale and those that focus on explaining single data sample at a time. Although local explanations might be initially sufficient for clinical applications as assistive diagnosis systems, global-level explanations will be crucial for understanding the model behaviour as a whole. This is specifically important for identifying decision biases and hence for the development towards autonomous decision systems. There exist various taxonomies for xAI methods in literature. We present a few types of methods that are specifically relevant to medical imaging. A visual overview of the grouping is provided in Figure \ref{fig:topology}.

\subsection{Visual Relevance Heatmaps}
Probably the most popular group of methods for explaining and interpreting image-based classification methods is the generation of visual heatmaps representing the influence of individual pixels on the result of the classification. Existing methods differ significantly in the computation of relevance values. The most obvious approach is the visualisation of the internal activations of a model~\cite{zeiler2014visualizing}. Therefore, single or combinations of intermediate, two-dimensional activation values are scaled to input size and visualised. Other common methods rely on attribution of the classification results to the individual pixels. In practice, this is done using e.g. weighted activations in Class Activation Mapping (CAM)~\cite{zhou2016learning}, gradient-based methods like Saliency~\cite{simonyan2013deep}, Gradient*Input~\cite{shrikumar2016not}, Grad-CAM~\cite{selvaraju2017gradcam}, Integrated Gradient~\cite{sundararajan2017axiomatic}, DeepLift~\cite{shrikumar2017learning} or methods based on mathematical decomposition like Layerwise-Relevance Propagation (LRP)~\cite{bach2015pixel}, Agglomerative  Contextual Decomposition (ACD)~\cite{singh2018hierarchical} and SHapley Additive exPlanations (SHAP)~\cite{lundberg2017unified}. All these methods require access to the model parameters and thus an understanding of the model architecture.

Perturbation-based methods, on the other hand, are completely model-agnostic and can thus be used for model-independent explanation without knowledge of their internals. In order to explain a given sample, it is modified several times and evaluated by the model again and again in order to systematically record the changes caused by the perturbations. Methods like Occlusion~\cite{zeiler2014visualizing}, RISE~\cite{petsiuk2018rise}, and Extremal Perturbation~\cite{fong2019understanding} differ in the occlusion strategy (procedure and perturbation). LIME~\cite{ribeiro2016should} goes one step further and trains local approximation models based on the results of the randomly modified images.

In addition to the post-hoc methods mentioned so far, there is also possibility to generate relevance heatmaps in an ante-hoc process. Here, model architectures can be extended by attention mechanisms that force the model to focus its attention explicitly on certain parts of the input and to hide the remaining part. This distribution of attention can often be visualised in a heatmap~\cite{ba2014multiple}, using pointers~\cite{jetley2018learn} or by explicitly cropping the input to the attended region~\cite{jaderberg2015spatial} to gain insight into the network’s decision-making process.

\subsection{Class- and Prototype Abstraction}
Visual Relevance Heatmaps (VRHs) usually help to explain the decision on individual samples. Another approach that aims towards both global and local explanations of DL models is the generalised representation of prototypes of individual classes or neurons as learned by the model. This includes, for instance, methods maximising the activation of particular outputs~\cite{simonyan2013deep} or intermediate neurons~\cite{mahendran2016visualizing} by optimising over an input image to determine their "prototypical" activation patterns. Many variations of this approach have already yielded interesting results and insights~\cite{olah2018building} for general image recognition tasks. However, only few works can be found applying abstraction methods to medical problems~\cite{graziani2018regression, graziani2019improved, toba2019quantitative, couteaux2019towards}. This might be due to the complexity and entanglement of disease criteria and consequently complications in interpreting the prototypical results.

\subsection{Conceptual Explainability and Biomarker Identification}
The aim of concept-based explanation methods is to map human understandable semantic concepts to the concepts learned by DL models after training in order to make their decision-making processes more comprehensible. Such concepts can be very simple characteristics such as colours, shapes or textures. However, complex concepts can also be defined, consisting of combinations of simpler concepts. The TCAV method developed by Kim et al.~\cite{kim2018tcav} requires only a small number of sample images per concept to compute global concept influence scores. Further exploitation of this method allows explicit localisation of the concepts recognised by the network in the input domain~\cite{lucieri2020explaining}, extend its application to regression tasks~\cite{graziani2018regression, graziani2019improved} and introduce improved metrics~\cite{yeche2019ubs}. Other concept-based approaches include the ones proposed by Bau et al.~\cite{bau2017network} and Zhou et al.~\cite{zhou2018interpretable}.

Especially in the application of DL in medical problems, the detection and localisation of biomarkers by the model is popular in addition to the diagnosis of diseases. This approach allows intermediate steps of the models to be validated by experts. As has been shown in recent works~\cite{graziani2018regression, graziani2019improved, lucieri2020interpretability}, even post-hoc concept-based methods can be used to detect such biomarkers. However, more common approaches in literature are ante-hoc methods based on multi-task learning~\cite{caruana1997multitask}, where the models are trained for the combined classification or localisation of biomarkers~\cite{kawahara2018seven, kawahara2018fully, coppola2020interpreting}. Segmentation networks are often used for localization as in~\cite{kawahara2018fully}, however, such explicit approaches presuppose that correspondingly annotated data are available. An alternative approach by Zhang et al.~\cite{zhang2019biomarker} combines the optimisation of a CNN and a Generative Adversarial Network (GAN)\footnote{Generative DNN that can be trained to learn underlying data generating process of a given training dataset. It can be used to interpolate between samples, generating unseen images.} in a single end-to-end architecture for the localisation of biomarkers without the presence of explicit biomarker annotations.

\subsection{Textual Explainability}
There are different methods for generating verbal explanations of DL model decisions. One can distinguish between methods that use a template approach~\cite{hendricks2018grounding,guo2018neural, munir2019tsxplain}, rule-based methods~\cite{srivastava2017joint,hancock2018training,lakkaraju2019faithful,rabold2019enriching} and methods that use Natural Language Processing (NLP) models to generate an explanatory text~\cite{hendricks2016generating,zhang2017mdnet}. An early use case of NLP-based, textual explanation generation in the medical domain is MDNet framework developed by Zhang et al.~\cite{zhang2017mdnet}. This framework allows to generate a textual diagnostic report based on a medical image. In addition, a heatmap is generated for each word of the diagnostic report, which shows users the model’s attention at that step.

\section{Achievements of xAI in Medicine}
\label{sec:achievements}

The number of research papers on interpretability and explainability of AI has mushroomed in the last few years~\cite{arrieta2020explainable} and thereby the application and adaption of xAI methods to specific medical domains have also increased. In the following, we present research with the most practical significance towards clinical decision support systems.

\subsection{Interventional Methods}
\begin{figure}[t!]
    \centering
    \includegraphics[width=0.75\textwidth]{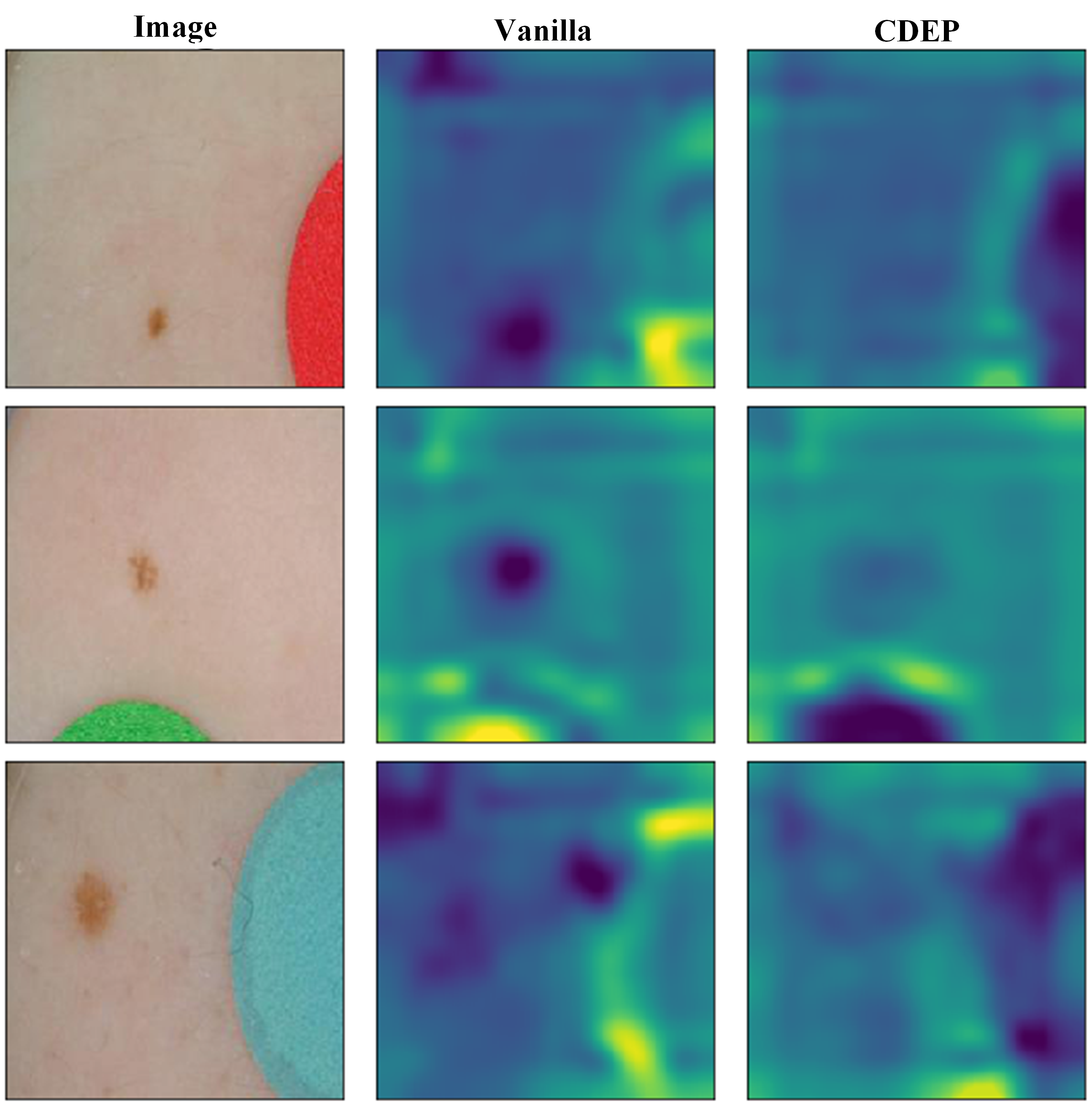}
    \caption{Results from Rieger et al.~\cite{rieger2019interpretations}. Left column shows original image samples from the dataset. The middle and right columns show grad-CAM heatmaps before and after correction using the developed model-correction method.}
    \label{fig:gradCAM}
\end{figure}

The explanation of high performing AI algorithms that utilise spurious indicators for classification allows to reveal those biases. To make practical use of these explanations, methods that facilitate intervention and correction of current working of algorithms are required. Common methods for penalisation and correction of explanations in DL models work by imposing a loss on explanation heatmaps (e.g. from VRH method) or conceptual predictions (e.g. TCAV) against ground truth explanations provided by human experts~\cite{ross2017right,erion2019learning}. This area is strongly related to the field of explicit expert knowledge incorporation. Examples of successful application of such methods in the medical domain are disease grading in diabetic retinopathy~\cite{mitsuhara2019embedding}, lymph node histopathology~\cite{graziani2020guiding} and dermoscopic skin lesion classification~\cite{yan2019melanoma,rieger2019interpretations}. Rieger et al.~\cite{rieger2019interpretations}, for instance, were able to correct a classifier trained on the ISIC 2019 dataset, which is heavily biased towards benign predictions when coloured patches appear beside the lesion. A comparison between Grad-CAM maps generated before and after correction of the network can be seen in Figure \ref{fig:gradCAM}. Inspired by the concept-based explanation method of TCAV, Graziani et al.~\cite{graziani2020guiding} fine-tuned a deep classifier for histopathologic lymph node tumour detection. By penalising undesired control targets (concepts), they managed to increase Area Under the Curve (AUC) by 2\%.

\subsection{Revealing new criteria}
Explainability methods are often employed in specific, sometimes medical application areas by expert computer scientist to prove their effectiveness. Lack of domain knowledge often hampers proper interpretation of presented results, rendering the provided explanations useless for assessing the correctness of the network. However, an increasing trend of collaboration between medical professionals and computer scientists is apparent in the application and tuning of DL models. The need for domain knowledge in order to understand and explain models has reflected in a growing number of publications on xAI including both computer scientists and domain experts.

A team of computer scientists and neurosurgeons succeeded in training a CNN for the localization of diagnostic features in confocal laser endomicroscopy images for glioma detection using only image-level annotations. Izadyyazdanabadi et al.~\cite{izadyyazdanabadi2018weakly} sequentially applied a visual relevance localization method (CAM) to a multi-head network, merging the resulting maps by collateral integration as well as biologically inspired lateral inhibition principle. Their diagnostic localization maps correctly identified familiar diagnostic features and also revealed new diagnostic regions that were previously unknown to the neurosurgeons.

Using a complex model architecture consisting of two autoencoders and further processing steps, an interdisciplinary team of pathologists and computer scientists successfully predicted the recurrence of prostate cancer from digitised slides of histological sections in~\cite{yamamoto2019automated}. An especially developed method for calculating an impact score, which provides information about the direction of influence of an image section for diagnosis, provides further insights. It has been confirmed that the model independently learned the concept of the Gleason Score, an established prognostic value for prostate cancer among experts worldwide, and has identified the occurrence of stroma, an intermediate tissue running through the parenchymatous organs, in areas of the incision free of cancer cells as a prognostic factor for prostate cancer.

No clear physiological characteristics of insomnia are known yet. Researchers from Charité Berlin, HTW Berlin and University Medicine Göttingen have used machine learning models in~\cite{jansen2019network} to detect insomnia in polysomnographic data with the aim of revealing such physiological features through AI. By applying DeepLift~\cite{shrikumar2017learning} method, some factors, such as increased and less synchronous eye movement, were highlighted as relevant for the prediction of insomnia. However, the authors themselves stress that the results should be interpreted with caution at first, as neither the bias of the results due to laboratory conditions can be excluded nor can the validity of the factors be definitively confirmed.

Lucieri et al.~\cite{lucieri2020interpretability} used the concept-based TCAV method to investigate a high-performing CNN for the classification of skin lesions into malignant melanoma, benign nevi and seborrheic keratosis. Using only a few particularly detailed annotated images, the authors were able to show that the CNN uses proven concepts defined by the medical community to classify images. These results were also confirmed by a team of experienced dermatologists. Part of the results regarding seborrheic keratosis, however, are yet to be confirmed.

A team of computer scientists and biologists have used samples of microbiomes of human female skin to determine phenotypes such as age, skin moisture, menopause status and smoking status in~\cite{carrieri2020explainable}. The SHAP method was used to assess the relevance of each bacterial genus in the microbiome. As this method generates local explanations, SHAP values for all bacterial genera were averaged over the subset of samples with correct and good results for classification and regression. The most relevant bacterial genera and their influence on the respective task were reported. For the determination of all phenotypes, a number of relevant bacteria genera were identified. In the case of skin moisture determination, for example, the genera identified by the model as particularly important were already associated with skin moisture in previous studies.

Essemlali et al.~\cite{essemlali2020understanding} were able to determine whether patients suffer from mild cognitive impairment or even Alzheimer's dementia using their two-dimensional connectivity matrix of brain regions. They used a specially adapted CNN architecture for this purpose. To explain the disease prognosis, the gradients of all images of the respective classes were averaged to obtain a global explanation. These averaged heatmaps of different classes were subtracted to emphasize the crucial differences between two conditions. The results confirmed that the connectivity of the entorhinal cortex is crucial for the separation between healthy and Alzheimer's disease subjects and the hippocampus for the separation between healthy subjects and those with mild cognitive impairment. Their results have been discussed with an expert neuroanatomist.

\subsection{xAI Frameworks}
\label{subsec:xai}
Since its rise in the early 2010s, DL and our understanding of its workings have greatly evolved. Thus, it is not surprising that the first commercial and non-commercial applications of explainable DL-based software solutions entered the market.

Data Language (UK) Ltd.~\cite{scopa2020UK} already use their commercial SCOPA Explainable AI Platform solution in classification of terrorist and extremist propaganda and claim it to be applicable to healthcare applications such as bone fracture detection from radiology images. The framework is described as using a layered ensemble of Deep Learning, Machine Learning and Algorithmic models for holistically explainable and scalable media classification.

The US-based start-up Decoded Health offers a commercial telehealth platform for radical automation of patient-doctor interactions and automated decision support for doctors~\cite{decodedHealth2020US}. Among other things, the company uses the Deep Adaptive Semantic Logic (DASL)~\cite{sikka2020deep} and ARSENAL~\cite{ghosh2016arsenal} methodologies developed by its parent company SRI International to explain DL models using rule-based reasoning and providing natural language interaction.

Explainable AI for Dermatology (exAID) is a showcase framework for computer-aided diagnosis of diseases developed by German Research Center for Artificial Intelligence GmbH (DFKI)~\cite{dfki2020exaid}. exAID is based on concept-based explanation methods~\cite{lucieri2020explaining,lucieri2020interpretability} to detect and localise biomarkers and generate verbal diagnostic explanations. In addition, the tool offers an advanced mode that allows data scientists and physicians in residence training to explore collected data sets in an exploratory way, including the suggestions of the DL model. The tool can be applied to any DL model and is thus compatible with all state-of-the-art architectures. 
%A demo of the system in example use-case of skin lesion classification is available on their website\footnote{https://exaid.kl.dfki.de/}.

Another DFKI project is specifically focused on the development of a Computer-Aided Diagnosis (CAD) system for the detection of skin diseases~\cite{nunnari2019cnn,sonntag2020skincare}. The system developed in the Skincare project is capable of analysing images of skin diseases taken with a smartphone, generating a differential diagnosis, and segmenting the skin lesion and individual biomarkers. The explainability of the system is ensured through the calculation of expert scores and VRHs. A demo of the system can be tested on the project webpage\footnote{http://www.dfki.de/skincare/classify.html}.

\section{Challenges for xAI Application}
\label{sec:challenges}

Since the initial applications of modern DL-based systems in medical domains, we have seen remarkable strides in the explanation of systems that in some cases already led to correction and verification of AI as well as disclosure of new potential diagnostic criteria. However, there are still a number of challenges pertinent to medical image diagnosis, which should be addressed by concerted efforts from AI researchers, medical practitioners and regulatory authorities.

\subsection{Evaluation of Explanation Methods}
Before xAI methods can be practically applied, it must be ensured that their explanations are reliable, trustworthy, and useful. In three steps, we distinguish between the evaluation of an explanation’s truthfulness, usefulness and finally the interpretation by the user.

\subsubsection{Evaluation of Truthfulness}
One of the key challenges in explainable AI is difficulty in evaluating if the explanation of a model’s behaviour is reliable, primarily because there is no ground truth available for evaluation~\cite{hooker2019benchmark}. Truthfulness or fidelity of an explanation refers to whether it is reliable and reflects the actual decision process of the AI. In order to practically deploy AI in clinical environments such that it increases the efficiency and accuracy of human doctors, it is of paramount importance to ensure the fidelity of xAI methods. However, due to lack of explanation ground truth, evaluation of such methods is largely subjective. 

There have been attempts to quantify and measure the quality of explanations. Samek et al.~\cite{samek2016evaluating} introduced Area Over the MoRF Perturbation Curve (AOPC) measure to quantitatively compare VRHs. The measure gradually perturbs input images starting from the regions that are marked as the most relevant according to a given explanation method. High AOPC values indicate that a model is sensitive to perturbations in those regions, thus confirming the validity. The RemOve And Retrain (ROAR) framework~\cite{hooker2019benchmark} is an advancement of AOPC approach. As image perturbations lead to a change in image distribution, they retrain the network on the perturbed images to avoid distribution gaps and evaluate the achieved accuracy. However, the evaluation of an altered model cannot give reliable insights into the sensitivity of the original model. In~\cite{tjoa2020quantifying} a synthetic dataset with ground truth explanation has been generated for easier xAI method evaluation. Adebayo et al.~\cite{adebayo2018sanity} introduced randomisation tests in which model weights and data labels where systematically randomised to reveal if explanation methods where really model and data dependent. Although this method has not been used to quantify fidelity, its results are certainly meaningful for evaluation.

Truthfulness is the basis for robust and useful xAI. Results from works like~\cite{adebayo2018sanity} showed that some methods produce convincing explanations that are worth no more than simple edge detectors. Eitel et al.~\cite{eitel2019testing} performed a quantitative comparison of visual relevance methods for MRI-based Alzheimer’s disease classification. They found that guided backpropagation attribution maps~\cite{springenberg2014striving} averaged over all true positives for multiple training runs highlighted different regions in brain MRI. However, despite the variance, which makes it harder to compare and to replicate outcomes of individual experiments, some regions like hippocampus, cerebellum and edges of the brain were commonly identified as salient regions. Other visual relevance methods like Gradient*Input, Occlusion Sensitivity, and LRP also showed similar behaviour, which raises serious questions on the robustness and coherence of these explanation methods. However, this could also indicate abundance of biomarkers in the data that allows DNN’s to perform the same task in a variety of ways.

\subsubsection{Evaluation of Usefulness}
Besides evaluating the fidelity and completeness of explanation methods, it is also crucial to quantify and qualify the usefulness of generated explanations. Doshi-Velez and Kim~\cite{doshi2017towards} proposed the distinction between application-grounded, human-grounded and functional-grounded evaluation of explanations. In~\cite{nguyen2020quantitative} the first functionally-grounded metrics where introduced, allowing to objectively judge the quality of an explanation. This quantification has the advantage of being independent from human subjectivity. On the other hand, human-grounded evaluation makes use of non-specialist human evaluators to subjectively compare or rate explanations. The evaluation approach that we find most important for xAI in medicine is the application-grounded evaluation. Depending on the domain or problem, medical practitioners have a very specific way of thinking of a problem, communicating or explaining a diagnosis. Hence, we argue that application-grounded evaluation is necessary to find and optimise the right explanation methods for a medical use-case.

\subsubsection{Evaluation with respect to Evaluators}
An equally decisive factor in the use of xAI methods is their interpretation by the end user. One and the same explanation can be interpreted differently by different individuals. A wrong or too naive interpretation of decision processes by developers or users can lead to serious consequences in the practical use of AI. The approach to the interpretation of explanations differs significantly for AI researchers and medical practitioners, but also overlaps to some extent.

For AI developers, explainability methods can help them design better models by understanding the interactions between the model and the data. However, AI developers and data scientists can sometimes over-trust or misuse these interpretability tools as noted by~\cite{kaur2020interpreting}. They conducted a small-scale study to learn how data scientists utilise publicly available interpretation tools and found that visual explanations are usually taken at their face values and used for rationalisation of suspicious observations instead of understanding how AI models worked. Experienced data scientists, on the other hand, were able to capitalise on these interpretability tools and effectively understand issues with models and data.

For medical practitioners, such tools can provide reasoning for model predictions and, therefore, develop trust and ease their acceptance into routine clinical workflow. Sayres et al.~\cite{sayres2019using} evaluated the impact of DL-based diabetic retinopathy (DR) detection algorithm on the performance of human graders in computer-assisted setting. They found that the accuracy of human graders improved when assisted by the algorithm that provided only disease prediction without any explanation. However, when the graders were provided prediction plus visual explanation by the algorithm, their detection accuracy improved only for patients who had diabetic retinopathy (resulting in high sensitivity) and decreased for patients without DR (resulting in low specificity). Although the qualitative feedback of human graders on the explanations provided by the algorithm was generally positive, the participants were not able to harness this additional information to notably improve their performance. This could partly be because the pathologic features of DR are very tiny in size, inconspicuous and occupy only a fraction of the whole images space.

To meet the challenges in the evaluation of xAI, special focus should be placed on the evaluation of realistic applicability of methods in clinical environment. This includes truthfulness, robustness, quality and the actual usefulness of the methods. Through such detailed analyses, the agreement between medical expert knowledge and the knowledge gained from the model and data can be validated and evaluated and, possibly, new knowledge can be gained. A further dimension that should not be neglected when evaluating xAI applications in healthcare is the ethical assessment of the impact on individuals and society. As shown in section \ref{subsec:xai}, there is an increasing commercial interest in explaining AI decisions. This requires development of regulatory measures that both take into account different needs of different individuals and user groups and are adaptable to the constantly evolving AI technology~\cite{tjoa2019survey}. However, this also requires clearly defined evaluation and certification processes to assess the ethical conformity of the use of AI in a specific context. z-Inspection~\cite{zInspection2020} is one of the first ethical evaluation and certification processes that integrates theoretical principles for the ethical evaluation of AI into a practically applicable framework. 

\subsection{Deployment in Clinical Workflow}
Proof of concept studies and prototype methods are required to be tested rigorously to analyse their contextual fit in real-world clinical environment. However, many obstacles have been discovered and highlighted by researchers in implementing laboratory research in clinical settings. These challenges include lack of utility to clinicians’ logistical hurdles that hamper clinical deployment and trials~\cite{elwyn2013many}. Ineffective use, or misuse, of these assistive systems can even lead to performance degradation of human graders~\cite{cole2014impact,kohli2018cad,cai2019human}. Cai et al.~\cite{cai2019human} developed interactive user-centric techniques for pathologists to improve diagnostic utility and trust in algorithmic predictions, in lab settings. Previously, such Human-Computer Interface (HCI) techniques have been used only to improve the algorithm, however, these interactive tools have potential to enable users to test, understand and grapple with AI algorithms, leading to new ways for improving explainability of algorithms. Instead of waiting for algorithms to generate human understandable explanations~\cite{cai2019effects,ribeiro2016should}, interactive techniques can allow users to play an active role in the interpretation of algorithm predictions and hypothesis-test their intuitions. While these studies help understand the needs of clinicians as they interact with AI algorithms, they do not account for the highly situated nature of activities in clinical environments. In a study~\cite{yang2019unremarkable} designed for field assessment of a Decision Support System (DSS) for cardiologists, it was found that the clinicians were more likely to embrace and use such systems if it was seamlessly and unobtrusively integrated into their existing workflow. However, the misuse of these systems can sometimes let the clinicians develop their own tolerance and workarounds in order to trust the algorithm results~\cite{jirotka2005collaboration}.

There are a few examples of such translation of AI into commercial applications for instance in detection of diabetic retinopathy~\cite{beede2020human}, cancer, and analysing radiology images~\cite{beam2016translating}. Deployment of CAD solutions in clinical settings can also help focus on the effects of workflow when new diagnostic and information systems are introduced into clinical environments. Arbabshirani et al.~\cite{arbabshirani2018advanced} integrated their AI based model for identification of Intracranial Haemorrhage (ICH) using head CT scans into clinical workflow for three months. During the trials, the model was able to reduce median time to diagnosis for routine studies from more than eight hours to only 19 minutes, while at the same time discovering some probable ICH cases which were overlooked by radiologists.

\subsection{Diverse and Complete Explanations}
Most applications of xAI in research focus on utilising single approaches and modalities for the explanation of AI models in given use-case. This can be seen in our analysis of achievements of xAI in section \ref{sec:achievements} as well as many reviews on xAI research~\cite{tjoa2020quantifying,stiglic2020interpretability,singh2020explainable,vilone2020explainable}. We believe that the integration of xAI in clinical workflows requires combination of multiple explanatory views to draw explanations that are diverse and as complete as possible. This is inspired by medical practitioners in routine clinical environments using textual descriptions alongside visualisations and temporal coherence to communicate decisions effectively and reliably. On one hand, this should motivate AI researchers to think of new, creative paths for xAI methods to complement existing methods and on the other to not only evaluate the effectiveness of approaches in isolation but in combination with diverse methods to leverage synergies. First efforts towards diverse explanations have been recently made in the Visual Question Answering community in works like~\cite{huk2018multimodal} and~\cite{alipour2020study}. Huk Park et al.~\cite{huk2018multimodal} show the positive complementary effect of visual relevance and textual explanations which is backed up by human evaluation. Completeness of explanations can be considered from a model’s and a user’s point of view. Completeness from a model’s point of view is directly related to fidelity. Yeh et al.~\cite{yeh2020completeness} introduced a completeness measure that quantifies the completeness of a given concept-based explanation for a model prediction. Completeness from a user’s point of view is subjective but equally relevant to usefulness.

\subsection{Human-Centric Explanations}
High-performing DL-based DNNs often utilise unintelligible notions of concepts to reach a prediction. Integration of AI assistants in clinical workflows requires human-centric explanation of decision that is able to not only explain a decision with high fidelity, but also conforms to human-understandable thought models. Compared to simpler use-cases like visual object classification or part segmentation, complex medical concepts used for diagnosis particularly necessitates to make explanations as human-understandable as possible.

\subsubsection{Human-Understandable Concepts}
One way to explain the decisions of AI based CAD systems in a human-centric way is to investigate the role of human-understandable concepts, learned by DL-based algorithms. It is very important to analyse the learned features of an algorithm that makes right decisions but based on wrong reasons. It is a major issue that can affect performance when the system is deployed in the real world. Explaining the role of a model’s concepts can reduce reliability concerns of medical practitioners and help develop their trust on CAD.

Application of concept-based xAI methods in medical image analysis has been problematic partly because these methods require concept datasets~\cite{zhou2018interpretable} or image patches corresponding to human-understandable concepts~\cite{kim2018tcav}, which are not always available. An unsupervised approach, extending CAV method, is developed by Ghorbani et al.~\cite{ghorbani2019towards} to cluster object datasets by performing segmentation of single objects and clustering their relevant activations into semantically meaningful groups. This approach cannot be directly applied to, for example, skin lesion classification where there is a substantial overlap between various concepts that can not be segmented into distinct spatial patches. Also, this method does not guarantee discovery of human-understandable concepts and requires thorough human evaluation effort.

Sometimes general explanation methods cannot be readily used for certain medical image tasks due to technical requirements or inappropriateness to the domain. Besides the continuous development of advanced xAI methods, it is crucial that developers pay attention to the domain specific needs of particular medical applications and their users. There have been many studies extending existing methods to better suit the challenges of the medical image analysis. For example, Yang et al.~\cite{yang2019weakly} proposed Expressive gradients (EG), an extension of commonly used Integrated Gradients~\cite{sundararajan2017axiomatic} to cover the retinal lesions better while~\cite{graziani2018regression} extended CAVs from~\cite{kim2018tcav} for continuous concepts like eccentricity and contrast. Lucieri et al.~\cite{lucieri2020explaining} extended the method for localising and highlighting image regions significant for network’s concept recognition in a medical inspired dataset. This could allow doctors to verify the network’s concept learning and suggest precise image regions for concepts. Such studies lead to the advancement of the xAI domain and provide specialisation to application domains without designing new methods from scratch. 

\subsubsection{Challenges in Textual Explanations}
Most disease classification algorithms using medical images attempt to answer Multiple Choice Questions in which the algorithm is expected to select one disease from a list of all possible diseases. In this type of experimental setting there is a fair chance that a correct prediction given by AI-based CAD is nothing more than a fluke – though the probability of fluke decreases with increase in total number of classes. Therefore, such classification algorithms require explicit interpretations of network predictions to validate their results.

In many medical domains like radiology and histopathology, doctors routinely write textual reports clearly noting salient findings before providing their impression (diagnosis). The nature of this type of detailed diagnosis substantiated by textual descriptions of the image is self-explanatory – at least for the domain experts. AI-based CAD can be enabled to process this multi-modal data (image and text) and generate textual reports to mimic the behaviour of radiologists and histopathologists. Such systems embed explanations of their decisions inside their predictions. These natural language explanations, using domain-specific terminology and mimicking structure of communication provide an intuitive and effective way of explaining decision processes to practitioners. However, providing textual explanations in the form of clinically accurate medical reports for medical images has some differences compared to other application areas where NLP is used to describe an image.

Generating long coherent reports (more than few tens of words) is one of the major challenges in textual xAI. Language generation models usually start with a few coherent sentences and after that its performance tapers off generating completely random words that have no association with the previously generated words or phrases. This happens generally due to very long temporal dependency among words which Long Short-Term Memory (LSTM)~\cite{hochreiter1997long} models have difficulty handling. One way to address this problem is to use transformer networks~\cite{vaswani2017attention} as language model decoder. These transformer models are able to capture relationship between words in a longer sentence better than Recurrent Neural Network (RNN) based models. Input text reports are tokenised and passed to the transformer network that consists of a single decoder layer and generates a query vector for another transformer model that generates reports by combining this query with information obtained from image processing model. The size of the generated reports and vocabulary can also be limited to ensure that the text is coherent and clinically meaningful.

Most of the reports written by doctors are free text report, which means that they don’t always follow any defined template. Reports written by two radiologists, for example, for a given X-ray image can be vastly different. There can be superfluous information that does not contribute directly to the final diagnosis. This makes it very difficult to compare AI-generated reports with human-generated reports especially when some of the reports depend on the previous chest X-ray of the patients and provide a continuous diagnosis based on previous examination. This problem can be addressed by removing those parts of the input reports which bear no influence on the diagnosis such as at what time the doctor saw the patient, or who was the doctor on call.

\subsubsection{Incorporation of Context}
Traditional AI algorithms overwhelmingly rely on one input modality, for example images in medical image analysis. However, medical practitioners routinely incorporate context, in the form of, for instance, patient’s clinical history, age, and sex etc., in their decision-making process. Compared to raw image pixels, this contextual information is much easier to understand for practitioners. Incorporation of this metadata into AI algorithms is tricky since context is difficult to represent in a form which is appropriate for processing by AI algorithms~\cite{cabitza2017unintended}. Not leveraging this useful context in AI algorithms can not only restrict their performance but also make explanations challenging. Therefore, another direction of research to make AI algorithms more transparent and explainable is to use multimodel data like medical images and patients’ records together in the decision-making process and attribute the model decisions to each of them~\cite{singh2020explainable}. This approach simulates the diagnostic workflow of a clinician where both images and physical parameters of a patient are used to make the decision. It can not only improve diagnostic performance of these algorithms but also explain the phenomena more comprehensively.

An interesting example in which context mattered and lack of its inclusion resulted in a technically valid yet misleading ML prognostic model was the use of mortality risk prediction to make decisions about whether to provide treatment on an inpatient or outpatient basis for more than 14000 patients with pneumonia~\cite{caruana2015intelligible}. In this study the algorithm counter-intuitively suggested that patients with pneumonia and asthma were at lower risk of death compared to patients with only pneumonia, an indication that surprised the researchers who eventually ruled it out. A closer analysis of the data revealed that, at the hospital hosting this study, patients with history of asthma who presented with pneumonia were usually admitted directly to intensive care units to prevent complications. This led to a pattern in the data that reflected better outcomes for such patients compared to patients with pneumonia and without history of asthma with approximately 50\% less mortality rate. This example not only emphasises the importance of representative training data for such algorithms, but also that a contextually complete description of the data is of crucial importance.

\section{Conclusion}
\label{sec:conclusion}

Medical Image Analysis has benefited a lot from recent advances in deep learning. However, beyond academic research and proof of concept studies, there has been a healthy scepticism about to what extent, if at all, AI should make or support medical decisions in real clinical workflows~\cite{holzinger2019causability}. Although the sequential computation of DL-based models is traceable, they often lack explicit declarative representation of knowledge. Justifying the decisions taken by AI using explanations can help bring such academic research one step closer to practical deployment in healthcare sector. We showed that the collaboration of AI developers and medical professionals already led to interesting advances in medical AI, including practical AI evaluation and discovery of new potential diagnostic criteria. It also appears that there is a growing interest in commercialising xAI solutions and developing use-case specific frameworks. However, we also caution to carefully gauge those achievements and continue investing efforts in standardised evaluation and investigation of xAI methods in close cooperation with domain experts.

In the medical domain, it is imperative to explain the output of algorithms in a human-understandable language as to support and not distract experts. Holzinger et al.~\cite{holzinger2017we} believe that the only way forward towards explainable CAD is to combine knowledge-driven and data-driven approaches, which could harness interpretability of former method and high accuracy of latter. We believe that the transition towards multimodal, diverse, and complete explanations that combine human-understandable modalities such as text, human-understandable concepts and context will substantially support the way of xAI in clinical assistive settings. In medical diagnosis, explanations can be different for different users. For instance, a doctor might use different language, modality or depth of explanation depending upon whether he/she is explaining to a patient, a regulator, or a fellow doctor. Similarly, explainable AI for healthcare serves different purpose for medical practitioners and AI developers. It is inevitable that AI engineers design solutions that provide diverse explanations fitting the need of specific use-cases. It has been observed that communication gap between AI developers and its users can lead to misuse of technology~\cite{cole2014impact,kohli2018cad}) and eventual performance degradation~\cite{sayres2019using}.

Qualitatively evaluating an explanation with regards to its interpretability and completeness can be substantially subjective. Recently, there have been many efforts to quantify and qualify xAI methods and their explanations in objective and subjective ways. However, there are no agreed-upon and standardised evaluation procedures for explanation methods that can guarantee fidelity and rate quality. Development of standardised and objective evaluation criteria can greatly help benchmark upcoming explainable CAD systems and is thus an extremely important requirement for application in routine clinical environments. Moreover, appropriate regulatory measures should provide an ethical framework for the application of AI in healthcare, which can ensure safety and transparency through standardised evaluation and certification procedures.

Since DL-based models are data driven, they suffer from limitations and biases inherent in the data. For example, some data might suggest that a cohort who took a certain drug recovered quickly compared to those who did not. DL models can detect this correlation easily. However, if the causality between drug and recovery is missing from the data, the models cannot propose an acceptable explanation of their decision either~\cite{pearl2018book}. Therefore, explainable AI and particularly CAD solutions will hugely benefit from a carefully curated dataset which incorporates the context and does not leave out any confounding variables. Such dataset curation can be achieved by concerted and close collaboration between medical practitioners and AI developers right from the onset. Moreover, the advancement of interventional methods for the correction of algorithms and incorporation of explicit expert knowledge could account for small retrospective adjustments during trial phases.

We believe that modern AI technology has the potential to revolutionise healthcare in innumerable ways and that xAI plays a crucial role in creating a solid foundation of understanding and improving its functionality. Current advancements show that a close collaboration of medical domain experts and computer scientists paired with persistent efforts of AI experts to advance and develop new methods will eventually lead to many practical applications which are just an anticipation of what will be possible in the future.

\bibliographystyle{unsrt}  
\bibliography{AIinHealthcare.bib}  %%% Remove comment to use the external .bib file (using bibtex).

\end{document}